\documentclass[dvipsnames,format=sigconf,anonymous=false,review=false, nonacm]{acmart}
\usepackage{lipsum}

\usepackage{caption}
\usepackage{subcaption}
\usepackage{texlogos}
\usepackage{graphicx}
\usepackage{tabularx}
\usepackage{pifont}
\AtBeginDocument{%
  \providecommand\BibTeX{{%
    \normalfont B\kern-0.5em{\scshape i\kern-0.25em b}\kern-0.8em\TeX}}}

\setcopyright{none}

\newcommand{\Cpp}{C\texttt{++}\ }
\newcommand{\mv}{\texttt{MvSR}}
\begin{document}

\title{Multi-view Symbolic Regression}

\author{Etienne Russeil}
\email{etienne.russeil@clermont.in2p3.fr}
\orcid{0000-0001-9923-2407}

\affiliation{%
  \institution{Université Clermont Auvergne, CNRS/IN2P3, LPC}
  \city{Clermont-Ferrand}
  \country{France}
  \postcode{F-63000}
}

\author{Fabrício Olivetti de França}
\email{folivetti@ufabc.edu.br}
\orcid{0000-0002-2741-8736}

\affiliation{%
  \institution{Universidade Federal do ABC, Center for Mathematics, Computing and Cognition}
  \city{Santo André, SP}
  \country{Brazil}
}

\author{Konstantin Malanchev}
\email{malanchev@cmu.edu}
\orcid{0000-0001-7179-7406}

\affiliation{%
  \institution{McWilliams Center for Cosmology, Department of Physics, Carnegie Mellon University}
  \city{Pittsburgh}
  \country{USA}
  \postcode{PA 15213}
}

\affiliation{%
  \institution{Department of Astronomy, University of Illinois at Urbana-Champaign}
  \city{Urbana}
  \country{USA}
  \postcode{IL 61801}
}

\author{Bogdan Burlacu}
\email{bogdan.burlacu@fh-hagenberg.at}
\orcid{0000-0001-8785-2959}

\affiliation{%
  \institution{Heuristic and Evolutionary Algorithms Laboratory, University of Applied Sciences Upper Austria}
  \city{Hagenberg}
  \country{Austria}
}

\author{Emille E. O. Ishida}
\email{emille.ishida@clermont.in2p3.fr}
\orcid{0000-0002-0406-076X}

\affiliation{%
  \institution{Université Clermont Auvergne, CNRS/IN2P3, LPC}
  \city{Clermont-Ferrand}
  \country{France}
  \postcode{F-63000}
}

\author{Marion Leroux}
\email{marion.rouby_leroux@uca.fr}
\orcid{0009-0004-5889-6652}

\author{Clément Michelin}
\email{clement.michelin@uca.fr}
\orcid{0000-0002-8924-1809}

\affiliation{%
  \institution{Université Clermont Auvergne, CNRS, Clermont Auvergne INP, ICCF}
  \city{Clermont-Ferrand}
  \country{France}
  \postcode{F-63000}
}

\author{Guillaume Moinard}
\email{guillaume.moinard@lip6.fr}
\orcid{0009-0002-7208-0874}

\affiliation{%
  \institution{Sorbonne Université, CNRS, LIP6}
  \city{Paris}
  \country{France}
  \postcode{F-75005}
}

\author{Emmanuel Gangler}
\email{Emmanuel.Gangler@clermont.in2p3.fr}
\orcid{0000-0001-6728-1423}

\affiliation{%
  \institution{Université Clermont Auvergne, CNRS/IN2P3, LPC}
  \city{Clermont-Ferrand}
  \country{France}
  \postcode{F-63000}
}

\renewcommand{\shortauthors}{Russeil et al.}

\begin{abstract}
Symbolic regression (SR) searches for analytical expressions representing the relationship between a set of explanatory and response variables. Current SR methods assume a single dataset extracted from a single experiment. Nevertheless, frequently, the researcher is confronted with multiple sets of results obtained from experiments conducted with different set-ups. Traditional SR methods may fail to find the underlying expression since the parameters of each experiment can be different. In this work we present Multi-View Symbolic Regression (\mv), which takes into account multiple datasets simultaneously, mimicking experimental environments, and outputs a general parametric solution. This approach fits the evaluated expression to each independent dataset and returns a parametric family of functions $f(x; \theta)$ simultaneously capable of accurately fitting all datasets. We demonstrate the effectiveness of \mv\  using data generated from known expressions, as well as  real-world data from astronomy, chemistry and economy, for which an \emph{a priori} analytical expression is not available. Results show that \mv\ obtains the correct expression more frequently and is robust to hyperparameters change. In real-world data, it is able to grasp the group behavior, recovering known expressions from the literature as well as promising alternatives, thus enabling the use SR to a large range of experimental scenarios.

\end{abstract}





\title{Multi-View Symbolic Regression}

\maketitle

\section{Introduction}
\label{sec:Intro}

The core of our modern scientific knowledge is based on careful production and analysis of experimental data. In a traditional  scenario, the researcher's task is to give meaning to measurable results, and analyze them in the light of a hypothesis. Frequently, the goal of this exercise is to find a mathematical description which can, at the same time, describe the recorded outcomes according to the state of the art of the field, and predict results from future similar experiments. Scientists became experts in the highly non-linear thought process required to interpret and translate scientific knowledge into suitable mathematical expressions. Due to the increase in data complexity, this task has also been approached through a machine learning perspective with to automate the thought process.

Symbolic Regression (SR), for example, produces a mathematical expression with one or more variables that optimally fits a given data set. It searches for a parametric model $f(x; \theta)$ that minimizes a loss function measuring the goodness-of-fit to the input data. Primarily, it searches for the correct functional structure $f(x; .)$ and, within this process, also tries to establish the correct values of the parameters $\theta$. This method has been successfully applied to simulated scientific data sets in 
physics~\cite{cranmer2020discovering}, chemistry~\cite{hernandez2019fast}, medicine~\cite{lacavaFlexibleSymbolicRegression2023a} and social sciences~\cite{de2023understanding}, to cite a few. Traditionally, SR algorithms used ephemeral random constants to set these parameters values, but more recently, the algorithms apply a nonlinear optimization to fit the function to the data more accurately.

In many real scientific applications, the researcher is faced with different sources of data describing the same model but acquired from different viewpoints. In practice, this means that we have more data available to fit the model but, though they share the same functional structure, they may differ in parameter values. For example, consider data related to the contagion dynamics of a new virus collected from different populations. Even though the dynamics of the virus should be the same among all of the populations, the different aspects of a given population may influence the observed behavior in such a manner that all populations behave following a specific $f(x; \theta)$, but with different values of $\theta$. We argue that even if data of a single viewpoint (hereafter single-view) is abundant, multiple views can help to constrain the search space of hypothesis.

In this paper, we introduce the idea of Multi-View Symbolic Regression (\mv) that allows the practitioner to use the information from multiple sources and control the desired number of parameters to guide the search process to a parametric model that contains the correct number of parameters so it is neither too flexible (i.e., a universal approximator) nor too rigid (i.e., fitting only a single data source). This stimulates the discovery of scientific models that facilitates the analysis and interpretation of the phenomena. Specifically, in this paper, we propose an adaptation to the Operon~\cite{operon} package to calculate the fitness of a candidate solution by fitting this solution and calculating the base fitness individually on each data source and then returning an aggregated fitness of all sources. The experiments bring evidence that such procedure can successfully retrieve models capable of fitting all the provided datasets.

Section~\ref{sec:description} describes the algorithm and our adaptation to the fitness calculation. In Section~\ref{sec:exp} we describe the experimental methods used to assess the validity of our approach. Section~\ref{sec:results} show the obtained results followed by a discussion.  In Section \ref{sec:real_data} we apply \mv\ to real scientific datasets from different fields and discuss the functional forms proposed. Finally, in Section~\ref{sec:conclu} we conclude this paper with some final thoughts and future perspectives.

\section{Multi-View Symbolic Regression}
\label{sec:description}

The goal of Multi-View Symbolic Regression (\mv) is to search for a general parametric model that can simultaneously describe multiple data sets generated by the same underlying mechanism.

To illustrate this concept, imagine that we have data from multiple epidemic events from different countries caused by the same pandemic. If we aggregate these data into a single dataset and fit a regression model, we will obtain a model that returns the expected contamination for a given time $t$. This expected value is calculated over the sum of all events and can be seen as explaining the average contagion event worldwide. On the other hand, if we have a model $f(t; \theta)$ fitted independently for each event, the parameters values will reveal different aspects for each of them. For example, fitting this model to data from different countries will describe different reproduction rates and thus inform about the characteristics of local populations (i.e., widespread use of mask, density of the population, etc.). 

In summary, \mv\ finds a model $f(x; \theta)$ minimizing the aggregated error when fitting the parameters $\theta$ independently for each experiment, while applying a constraint on the number of parameters. More formally, given a dataset $\mathcal{D} = \left\{(x_i, y_i)\right\}_{i=1}^{p}$ with $p$ data points $x_i \in \mathcal{R}^{d}, y_i \in \mathcal{R}$, \textit{SR} seeks a parametric model $f(x; \theta)$ that minimizes a loss function $\mathcal{L}$ given the optimal parameter values $\theta$:

\begin{equation*}
    \min_{f, \theta} \mathcal{L}(f(x; \theta), y).
\end{equation*}

However, with \mv\, the objective becomes:

\begin{equation}
    \min_{f}{agg_{i=1 .. k}{\left(\min_{\theta^i}\mathcal{L}(f(x^i; \theta^i), y^i\right)}},
    \label{eq:mvsrfit}
\end{equation}
where the superscript $i$ refers to the index of each dataset, $agg$ is an aggregation function such as $max, avg, med$. Additionally, we impose the constraints

\begin{align*}
    x \in \mathbb{R}^m \\ 
    \theta \in \mathbb{R}^n \\
    \text{s.t., } n \in \mathbb{N}, l < n < u,
\end{align*}
where $m$ is the number of independent variables ($x$), $n$ is the number of parameters in the model ($\theta$) with $n$ being bounded by a finite subset of the natural numbers.

Overall, given $k$ different datasets, we want to find the function $f$ with a limited number of parameters that minimizes the aggregated value of $\mathcal{L}$ for each dataset when independently adjusting the value of $\theta$ for each set.
The purpose for these constraints is to find a model that can correctly adjust to the data without underfitting nor overfitting while containing the smallest number of parameters that summarizes each data set. A complete and ideal implementation of \mv\ should allow to:

\begin{enumerate}
    \item Control the maximum number of parameters.
    \item Allow reuse of parameters in the symbolic expression (i.e., $\theta_0$ can appear multiple times).
    \textbf{\item Receive multiple datasets as input.}
    \textbf{\item Optimize the parameters independently for each dataset.}
    \textbf{\item Use an aggregation function to compute an overall loss.}
    \item Penalize solutions based on the number of parameters used.
\end{enumerate}

For this work, we present a proof of concept of the ideal \mv\ approach where only the points \textbf{3}, \textbf{4} and \textbf{5} of the list above have been fulfilled. Our current implementation (detailed in Section \ref{sec:implementation}) slightly modifies a pre-existing Symbolic Regression algorithm at the evaluation 
phase. So, when a solution is evaluated, for each one of the $k$ datasets the algorithm will: fit the model to the datasets, calculate the losses, and then aggregate the results with the aggregation function. The algorithm will then use this result as a metric for reproduction (within the context of genetic programming).

A related approach was proposed in~\cite{factorSR} focused on a better way to deal with nominal variables. In this work, they create a new terminal symbol called \emph{factor variable} representing a numerical parameter that assumes different values depending on the value of the nominal variable. For example, if the dataset has a variable named color assuming the values $\{\text{red}, \text{yellow}\}$, each factor variable would hold two numerical values corresponding to each nominal value. Different from \mv\, they optimize all parameters values in a single optimization problem, reformulating the mean squared error by introducing factor variables (one-hot encoded from the nominal variables) interacting with the numerical values. In our approach, we do not assume the existence of nominal variables, but rather samples from different populations, thus we need to treat them independently in the optimization process. As a result, our fitness value corresponds to the worst fitted sample, which could alleviate the effect of underrepresentation of one of the populations.

\subsection{Implementation details}
\label{sec:implementation}

In this work we 
use Operon~\cite{operon}, a high-performance \Cpp framework supporting single and multi-objective GP with non-linear optimization of the parameters using the Levenberg-Marquardt algorithm ~\cite{kommenda2020parameter}. This framework was reported to perform well with respect to the runtime and overall quality of the results~\cite{srbench}. The framework also has a Python module counterpart, called PyOperon, which 
presents all the flexibility of the \Cpp version.

The implementation was adapted to support the creation of independent evaluators, each of which evaluates the symbolic expression to one of the inputted datasets. Each evaluator fits the expression to one dataset, such that the expression can have different adjusted parameters for each view,  evaluate the loss function, and then aggregate the losses using an \emph{aggregation function}. This value is used during selection and reproduction stages to calculate the probability of survival. 

The available aggregation functions are \emph{average, median, min, max, harmonic mean}. Notice that since Operon always minimizes the fitness, the \emph{min} and \emph{max} aggregation functions represent the best and worst fits, respectively. We use $max$ as the aggregate function, effectively making the objective as the function $f$ that minimizes the worst fit among the different sources of data. After the evolutionary process, the current implementation will return the symbolic expression fitted on the last dataset as a string. The best found expression is then converted into a parametric solution using SymPy~\cite{sympy}. The expression is simplified before replacing each float by a free parameter and converted into a python function. The particular float values of the last dataset are used as initial guess for future external minimization.

\section{Experiments}
\label{sec:exp}

To demonstrate the advantages of \mv\ when multiple data sources are available, we devised an experimental design using data artificially created from the same generating function with either different parameters or covering different regions of its domain. Besides artificial benchmarks, we applied \mv\ to three real-world datasets from different scientific fields and highlighted the benefit of using this approach instead of a traditional SR.

\subsection{Data generation}
\label{sec:data_gen}

For the artificial data, we set up a series of challenging benchmarks based on standards 
from the SR literature. For this purpose, we chose three generating functions:

\begin{align}
    f_1(x) &= \theta_0 + \theta_1 x + \theta_2 x^2 + \theta_3 x^3 \label{eq:f1}\\
    f_2(x) &= \sin{(\theta_0 x_0 x_1)} + \theta_1 (x_2 - \theta_2)^2 + \theta_3 x_3 + x_4\label{eq:f2}\\
    f_3(x) &= (\theta_0 x_0^{2} + (\theta_1 x_1 x_2  - \frac{\theta_2}{(\theta_3 x_1 x_3 + 1)})^{2})^{0.5} \label{eq:f3}
\end{align}
$f_1(x)$ is a third order polynomial function which constitutes a simple one dimensional case for which results can easily be visualized and interpreted. $f_2(x)$ and $f_3(x)$ are based on the Friedman functions~\cite{friedman} for which free parameters have been added by replacing some constant values. 

From the parametric functions, we generate examples which individually carries incomplete information about their parent function. For the polynomial function ($f_1$) we test two separate cases: i) each dataset is generated with two parameters set to $0$ (see Views 1 to 4 in Table \ref{tab:params}), ii) each dataset uses the same parameters (see partial view in Table \ref{tab:params}) but displays only a narrow part of the behavior from which extrapolation of the original function is challenging. For $f_2, f_3$ we will only test the first case where, by setting some the coefficients to $0$, we mischaracterize the original function with the goal of rebuilding it using the different views.

 \begin{table}[t!]
     \centering
     \caption{Parameter values used for each view and the partial view (only for $f_1$). For the different views we kept two parameters as $0$ to depict the extreme situation where some parameters have no effect into the data. 
     The partial has no parameters equal to $0$ but each example is very restricted in the sampling range.}
     \begin{tabular}{|c|c|c|c|c|c|}
         \hline
        & View 1 & View 2 & View 3 & View 4 & Partial view \\
        \hline
        $\theta_{0}$ & $2$ & $0$ & $0$ & $2$ & $2$\\
        \hline
        $\theta_{1}$ & $2$ & $2$ & $0$ & $0$ & $-2$\\
        \hline
        $\theta_{2}$ & $0$ & $2$ & $2$ & $0$ & $2$\\
        \hline
        $\theta_{3}$ & $0$ & $0$ & $2$ & $2$ & $2$\\
        \hline
     \end{tabular}
     \label{tab:params}
 \end{table}

Each of these benchmarks will contain a total of $4$ datasets (i.e., views) with the sample sizes fixed for each view.
$f_1$ datasets consists of 20 points equally spaced on the interval $[-2, 2]$. In the partial view case we keep the same number of points, but we sample each dataset into evenly spaced domains $[-2, -1], [-1, 0], [0, 1]$ and $[1, 2]$. $f_2$ and $f_3$ datasets consist of 100 points uniformly distributed on the intervals detailed in the scikit-learn package \footnote{\url{https://scikit-learn.org/stable/modules/generated/sklearn.datasets}}.

To homogenize the results among the different benchmarks, we scaled the target variable $y$ of each dataset by applying the transformation $y'_i = 10 \cdot \frac{y_i}{\max(\mathrm{abs}(y_i))}$. The factor 10 was arbitrarily chosen to multiply the absolute error, thus avoiding all scores to be clustered around 0 independently of the fit quality. This procedure introduces an extra scaling parameter for the $f_2$ and $f_3$ functions, increasing the true number of free parameters to $5$.

For each generative function, we create multiple datasets using different noise rates, 
$\left\{0.000, 0.033, 0.066, 0.100\right\}$, to verify the robustness of this approach w.r.t. to noise. The noisy target is sampled from the distribution $\mathcal{N}(y,  \sigma_y \sqrt{\frac{\tau}{1 - \tau}})$, where $\tau$ is the noise rate.

\subsection{Operon Hyperparameters and Post-processing}

As mentioned in Section~\ref{sec:implementation}, we used an adapted version of pyOperon~\footnote{\url{https://github.com/heal-research/pyoperon/releases}} supporting the use of multiple datasets and aggregation function. For the following experiments, we used the hyperparameters depicted in Table~\ref{tab:ghyper}.

Additionally, we varied the hyperparameter max tree size from $5$ to $25$ with increments of $2$. This experiment will serve two purposes: i) having a baseline of models simpler than the original generating function since the functions require a minimum size of $7, 11, 14$ to be correctly represented; ii) test whether \mv\ is prone to overfitting if given the freedom to expand the expression to larger sizes, thus providing opportunity to fit the noise term as well.

Each one of the $176$ combinations of functions (multiple views and, for $f_1$, partial domains), noise level and maximum size represents a single instance of our set of benchmarks.
For each instance we ran pyOperon main SR module for each one of the four views independently as well as with the \mv\ adaptation. Each experiment was repeated with $100$ different random seeds. After a run, the string representation of the best symbolic model is processed with Sympy to replace the numerical values with parameter variables and the corresponding expressions are stored for post-processing.

These expressions are individually refitted to the noiseless version of each dataset, minimizing the least squares with the python package iminuit \cite{iminuit}. The final score is calculated using the mean squared error (MSE). Therefore, if the correct expression (or any equivalent) is generated the score will be $0$, even for functions generated on noisy datasets. When larger trees are allowed, some runs may generate NaN values as output. In such case, the value is replaced by infinity.

\begin{table}
    \centering
     \caption{List of the fixed hyperparameters used in the experiments.}
    \begin{tabular}{|c|p{4.5cm}|}
    \hline
    \textbf{Parameter} & \textbf{Value} \\
       \hline
       population size & $1000$ \\
       \hline
       generations & $1000$\\
       \hline
       pool size & $5$ \\
       \hline
       error metric & $MSE$ \\
       \hline
       prob. cx / mut. & $1.0$ / $0.25$ \\
       \hline
       max depth. & $10$ \\
       \hline
       optim. iterations & $100$ \\
       \hline
       agg. function & max \\
       \hline
        operators&$+$, $-$, $\times$, $\div$, \verb!^! 2, exp, $\sqrt{}$, sin ($f_2$ only)\\
        \hline
    \end{tabular}
    \label{tab:ghyper}
\end{table}

\section{Artificial Benchmark Results}
\label{sec:results}

\begin{figure*}[t!]
    \centering
    \begin{subfigure}[b]{0.27\textwidth}
    \includegraphics[width=\textwidth]{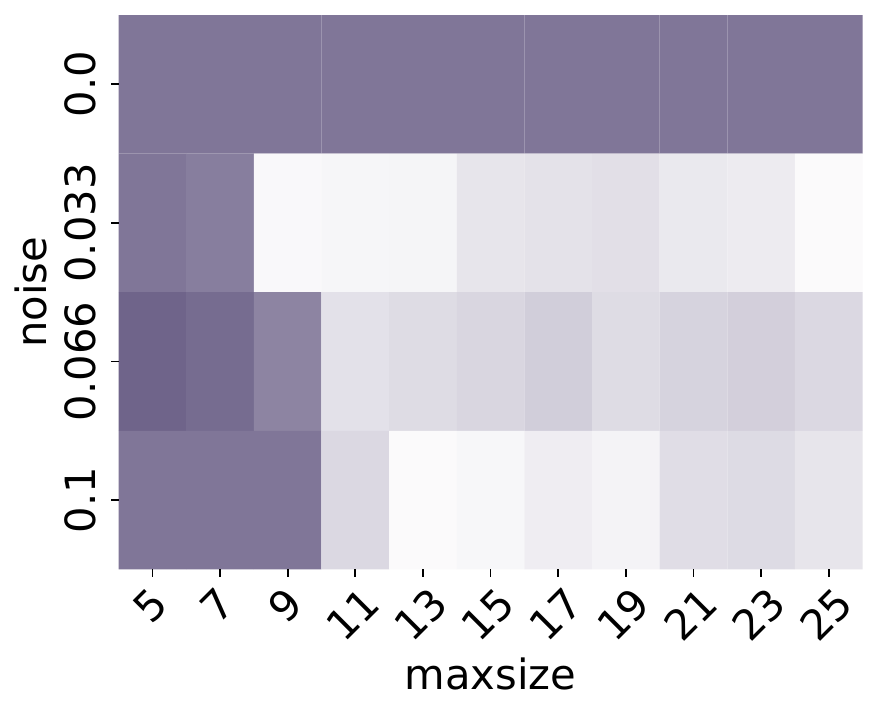}
    \caption{}
    \label{fig:poly-a}
    \end{subfigure}
    \begin{subfigure}[b]{0.27\textwidth}
    \includegraphics[width=\textwidth]{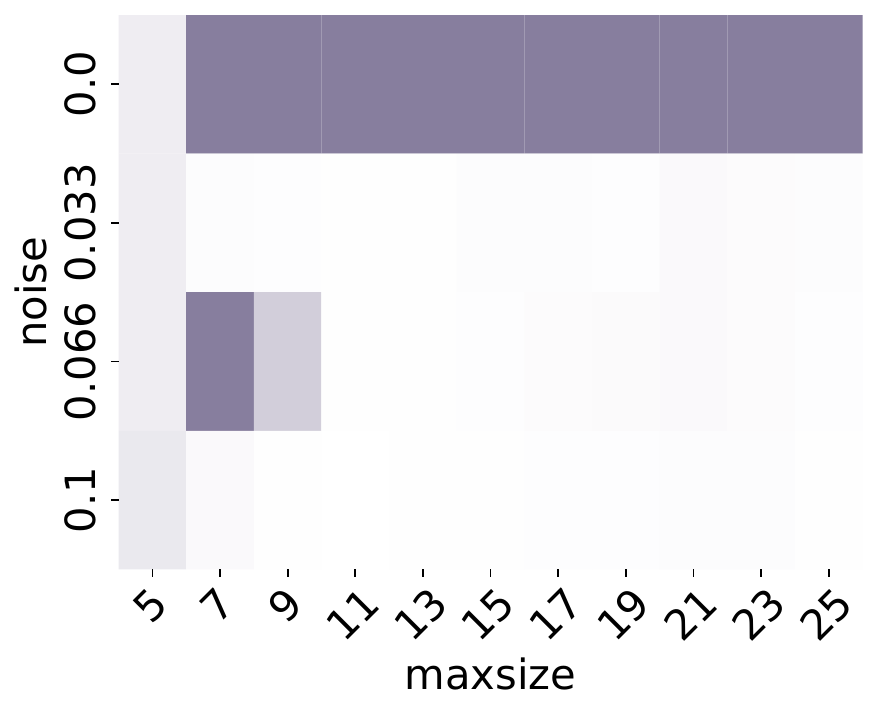}
    \caption{}
    \label{fig:poly-b}
    \end{subfigure}
    \begin{subfigure}[b]{0.27\textwidth}
    \includegraphics[width=\textwidth]{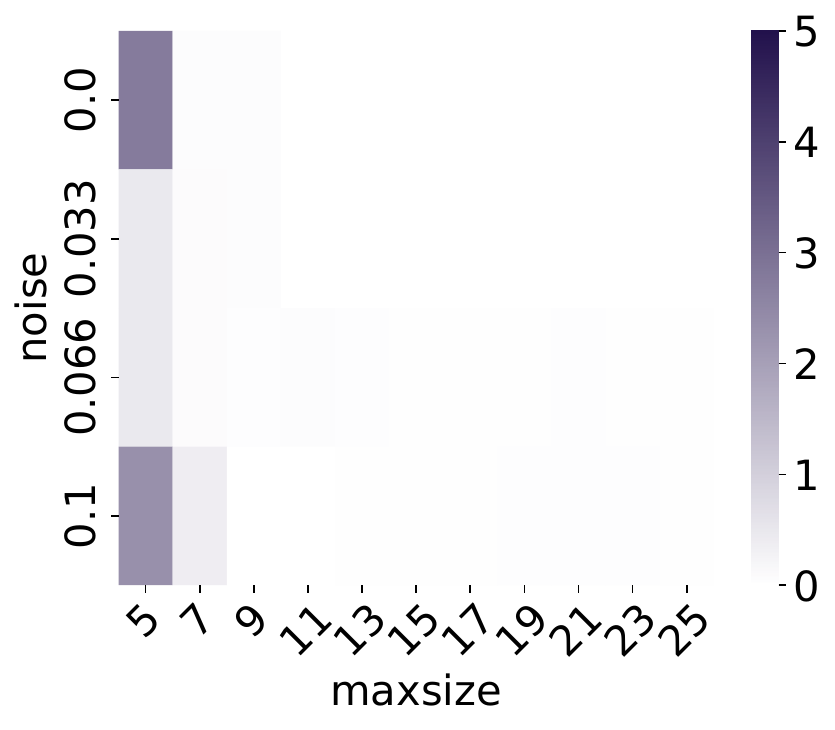}
    \caption{}
    \label{fig:poly-c}
    \end{subfigure}
    \begin{subfigure}[b]{0.27\textwidth}
    \includegraphics[width=\textwidth]{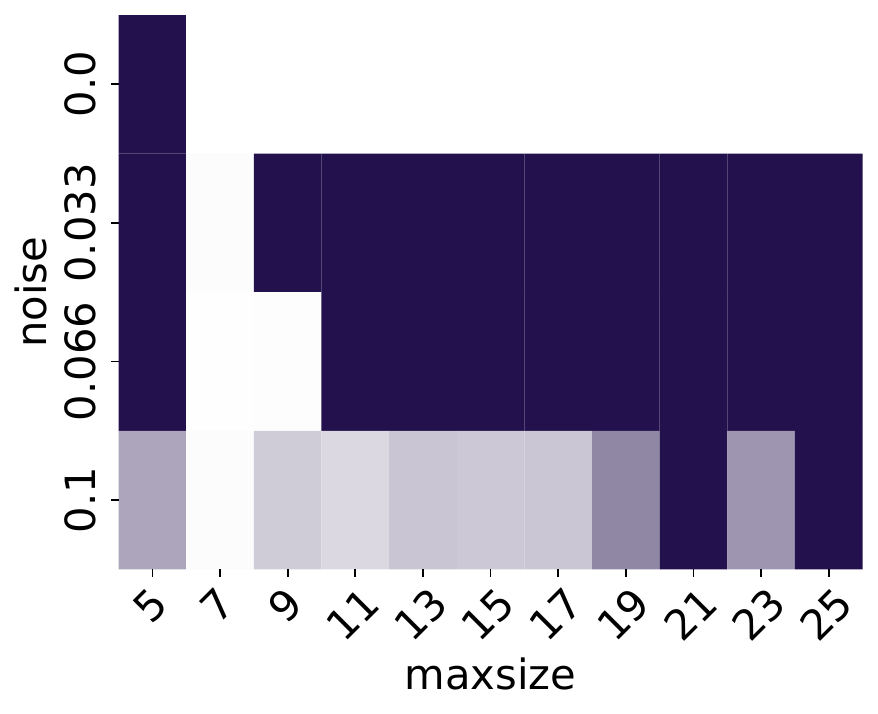}
    \caption{}
    \label{fig:poly-partial-a}
    \end{subfigure}
    \begin{subfigure}[b]{0.27\textwidth}
    \includegraphics[width=\textwidth]{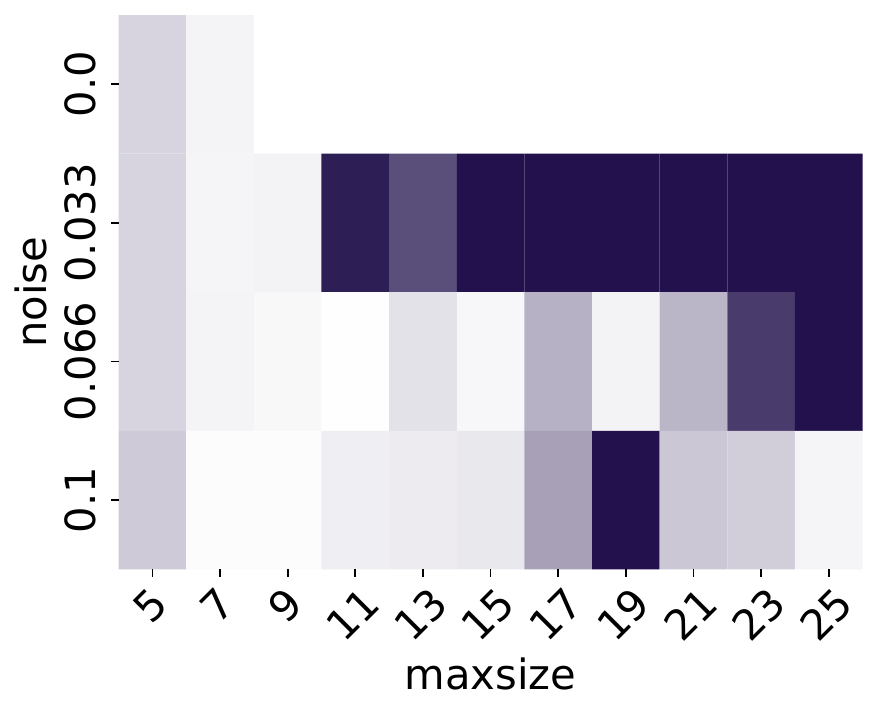}
    \caption{}
    \label{fig:poly-partial-b}
    \end{subfigure}
    \begin{subfigure}[b]{0.27\textwidth}
    \includegraphics[width=\textwidth]{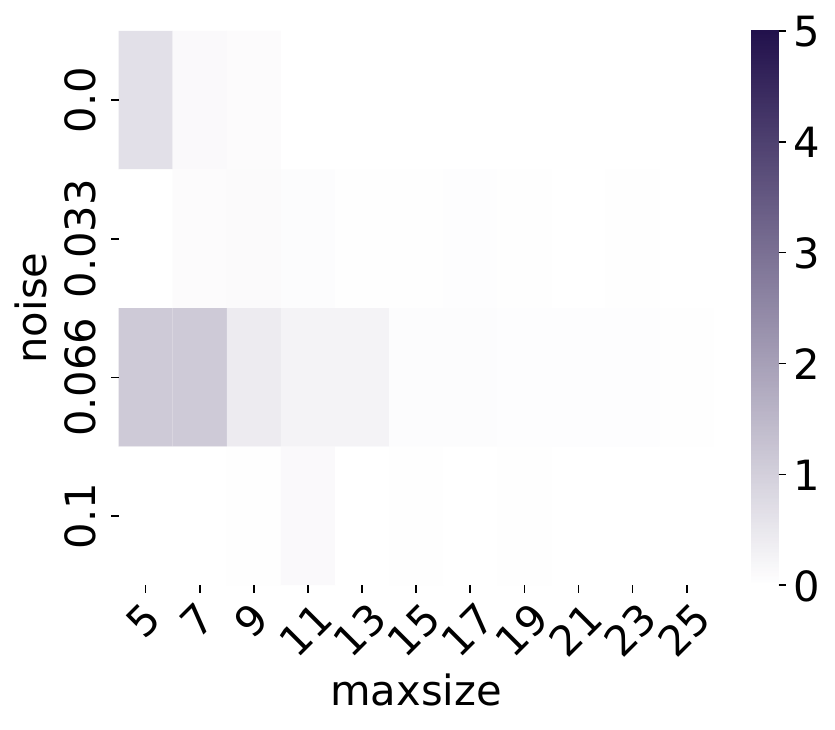}
    \caption{}
    \label{fig:poly-partial-c}
    \end{subfigure}    
    \begin{subfigure}[b]{0.27\textwidth}
    \includegraphics[width=\textwidth]{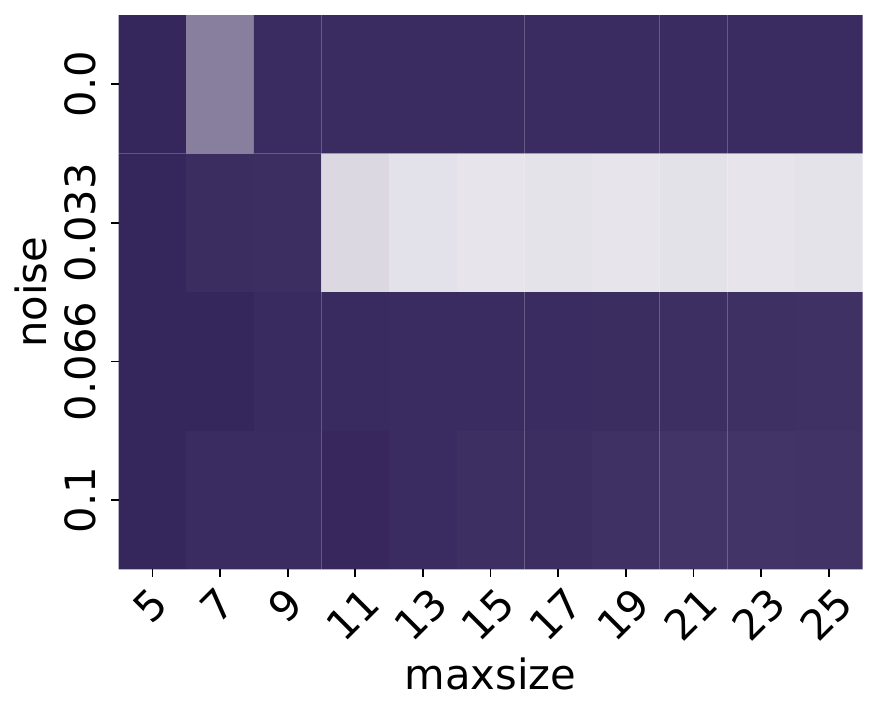}
    \caption{}
    \label{fig:friedman-a}
    \end{subfigure}
    \begin{subfigure}[b]{0.27\textwidth}
    \includegraphics[width=\textwidth]{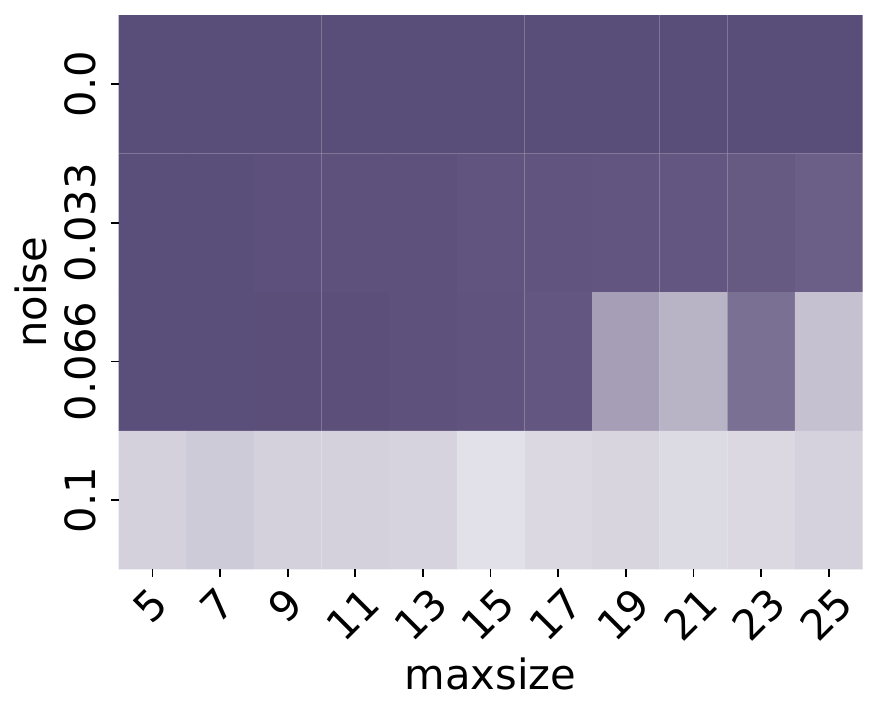}
    \caption{}
    \label{fig:friedman-b}
    \end{subfigure}
    \begin{subfigure}[b]{0.27\textwidth}
    \includegraphics[width=\textwidth]{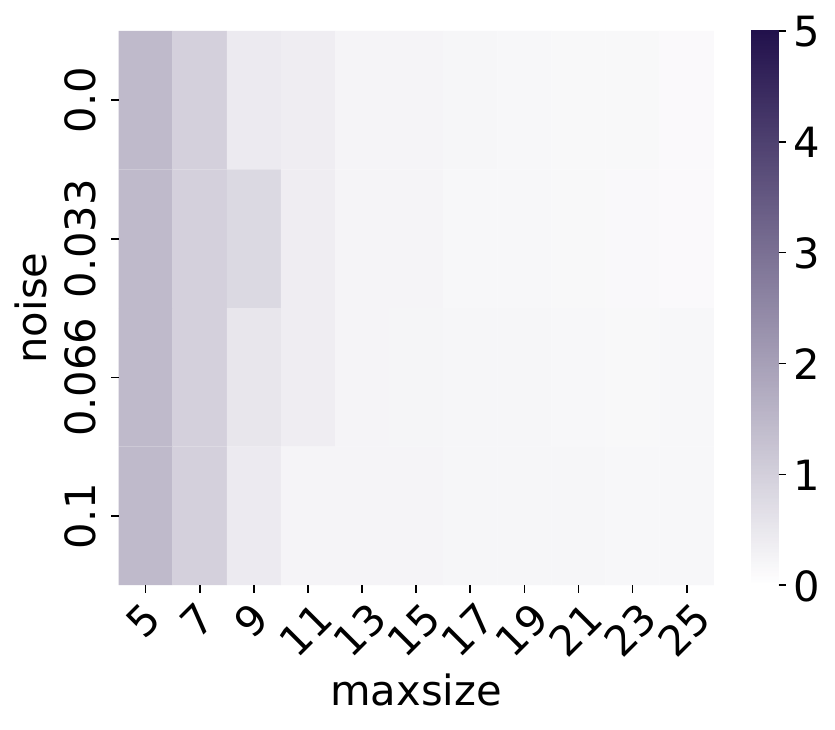}
    \caption{}
    \label{fig:friedman-c}
    \end{subfigure}    
    \begin{subfigure}[b]{0.27\textwidth}
    \includegraphics[width=\textwidth]{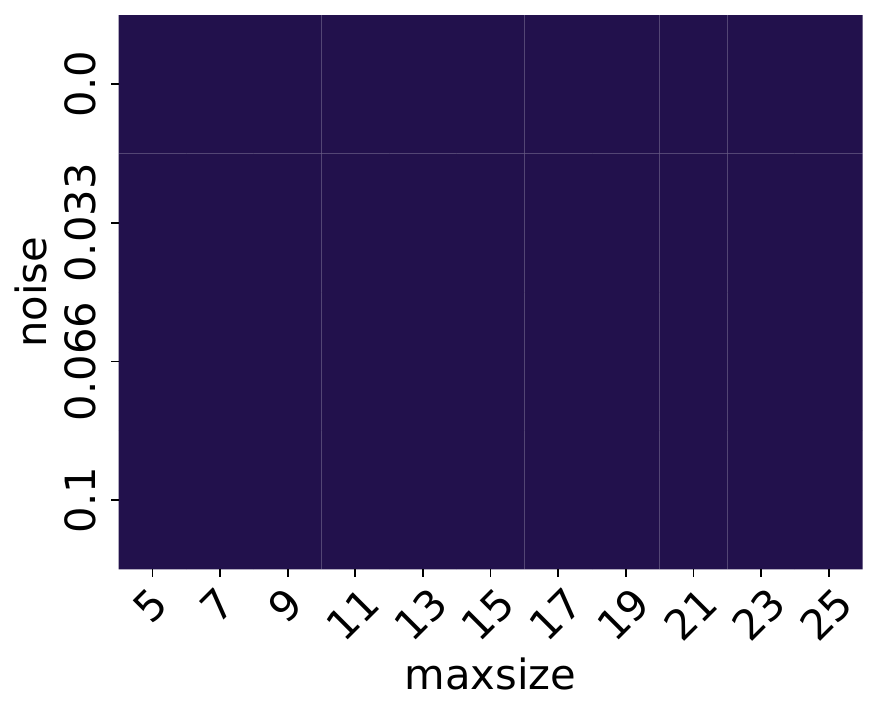}
    \caption{}
    \label{fig:friedman2-a}
    \end{subfigure}
    \begin{subfigure}[b]{0.27\textwidth}
    \includegraphics[width=\textwidth]{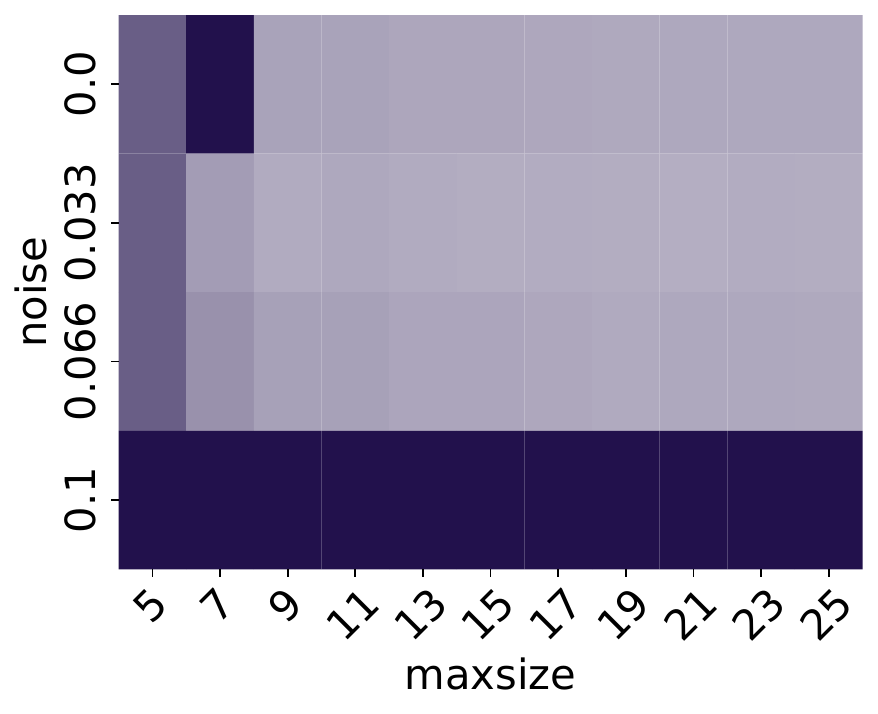}
    \caption{}
    \label{fig:friedman2-b}
    \end{subfigure}
    \begin{subfigure}[b]{0.27\textwidth}
    \includegraphics[width=\textwidth]{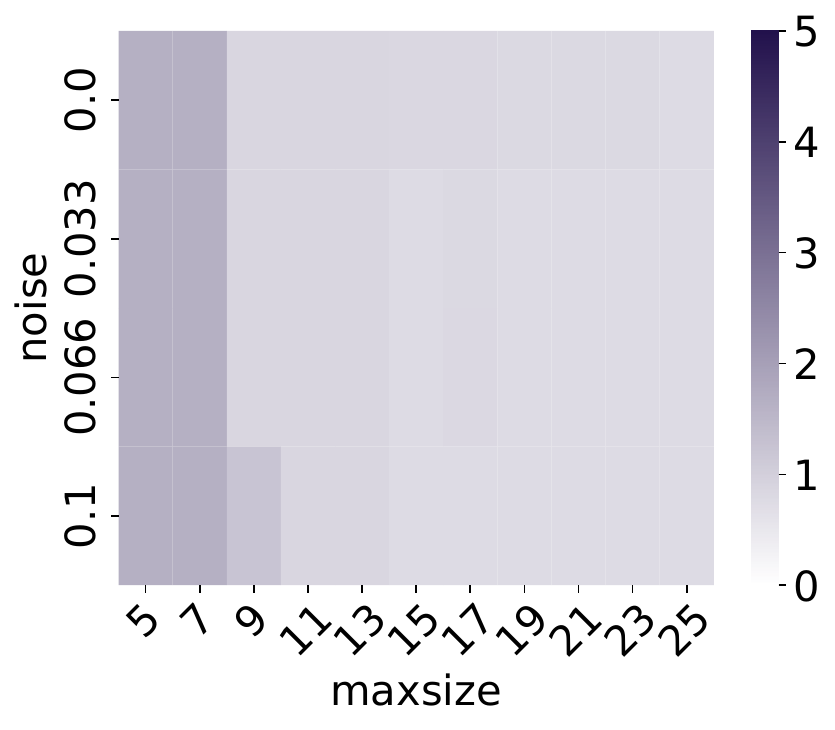}
    \caption{}
    \label{fig:friedman2-c}
    \end{subfigure}    
    \caption{Heatmap of the median MSE of the tested combinations of noise and maximum expression size. Each row of heatmaps show results for the $f_1$, $f_1$ partial domains, $f_2$ and $f_3$ benchmarks, respectively. Columns represent the worst single-view (left), best single-view (center), and \mv\ results (right). The colorbar represents the median MSE for that configuration, ranging from $0$ (white) to a clipped value of $5$ dark blue. The clipping improves the comparison of small values.}
    \label{fig:poly}
\end{figure*}

Figures~\ref{fig:poly} show heatmap plots of the results for each benchmark function for every combination of noise level and maximum size. The colorscheme displays the median MSE of each combination, the lighter the color the better the result. 
Any values higher than 5 (including infinity) is depicted as the darkest color in order to keep the contrast in the visualization of the results. 
In each plot, we show the worst and best single-view results (using the median of the plotted values as a choice criterion) and the \mv\ results.

Figures~\ref{fig:poly-a}~to~\ref{fig:poly-c} show that for the noisy datasets and after a certain value of maximum size, both the single-view and \mv\ behaves similarly reaching the correct solution. For the noiseless dataset, only \mv\ is capable of finding the correct solution. This is expected since the single-view approach is submitted to a simplified version of the original expression and chooses the simplest solution.

In Figures~\ref{fig:poly-partial-a}, \ref{fig:poly-partial-b} and \ref{fig:poly-partial-c} we can see a degradation of the results when using single-view with the noisy data. As the single-view has access to only one part of the polynomial curve, it can often fit the training data with a lower degree polynomial. This does not happen with the \mv\ as it recovers a correct solution in almost every combination of noise and maximum size. The same behavior is displayed in Figures~\ref{fig:friedman-a}~to~\ref{fig:friedman2-c} where, in every combination, \mv\ returns an optimal model (or near-optimal if the optimal model is not allowed by the maximum tree size) while the single-view approaches struggle with the lack of information and higher noise levels. We observe that \mv\ is resilient against noise and successfully prevents overfitting even when the maximum size is larger than the original expression.

Another important question is whether \mv\ returns a model with the correct number of parameters. A model with less than the optimal number of parameters is less flexible to fit the data from different sources and, thus, has a tendency to underfit. Similarly, a model with more parameters than needed, can be too flexible and capable of fitting the noise imbued in the data, thus creating an overfitted model. 

\begin{figure}[t!]
    \centering
    \includegraphics[width=0.4\textwidth]{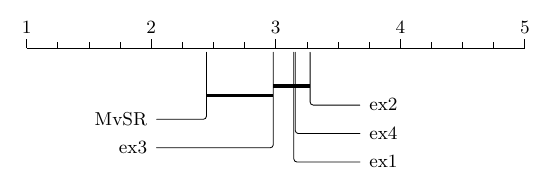}
    \caption{Critical difference diagram of the average rank w.r.t. the absolute difference between the number of parameters of a benchmark and the number of parameters of the model. This diagram was built calculating the Friedman hypothesis test with $\alpha=0.05$ and Holm–Bonferroni correction. The names \emph{ex1, ex2, ex3, ex4} refers to the four single-view models.}
    \label{fig:cd}
\end{figure}

Having the correct number of parameters is important when building a model to not only avoid under and overfitting, but also enable model interpretation. Figure~\ref{fig:cd} shows the critical difference diagram of the average rank calculated by comparing the absolute value between the optimal number of parameters and the number of parameters of each model. If the absolute difference is $0$ it means the model contains the exact number of parameters, anything higher than that means that it either has more or less than the reference number. This plot shows that \mv\ average rank (approx. $2$) is better than the single-view approaches. While the single-view approaches are indiscernible between each other.

\section{Scientific application}
\label{sec:real_data}

In this section we apply \mv\ to real experimental datasets coming from three scientific fields, namely chemistry, finance, and astrophysics. Such problems represent a significantly harder challenge for SR algorithms as the data generally have no absolute ``correct'' generative model, are irregularly sampled and display non-Gaussian noise. However, it constitutes perfect testing grounds for \mv, since one cannot know \textit{a priori} if examples provide display the full or partial behavior of the best parametric model. Therefore, in such cases, the most conservative approach is to always use \mv\ in order to build  general laws. 

\subsection{Chemistry dataset}
\label{sec:chem}

Beer’s law is an empirical law widely used in chemistry which relates the attenuation of light to the properties of the material through which the light is travelling. In chemistry, the attenuation of a beam of light going through a solution is presumed to be only due to absorption, as solutions do not scatter light of wavelengths frequently used in analytical spectroscopy. In UV-visible spectroscopy, we characterize a solution by its transmittance. This corresponds to the ratio of light intensities before and after passing through the sample. We express the absorption of a solution as $A= -\log(T)$. We experimentally observe that all molecules display a common pattern: for $A \leq 1$, the absorption rises linearly with the concentration, while for $A > 1$ the linearity breaks and the absorption increases more and more slowly until it reaches a plateau due to the limit of measurements of the spectrophotometer.

The Beer’s law is used to describe the properties of chemical species when $A \leq 1$. It states that the absorptive capacity of a dissolved substance is directly proportional to its concentration in a solution and is expressed as $A = \epsilon lc$ where A is the absorbance, $\epsilon$ is the molar extinction coefficient, l is the optical path length and c the concentration. A handful of alternative models have been proposed. They attempt to build a general Beer's law capable of fitting a larger range of absorption. For example, \citet{beer_poly} suggested using a quadratic polynomial equation to compensate the positive or negative deviations from the linearity. \citet{BLB_small} proposed to extend law by adding two exponent parameters on $l$ and $c$, thus adding flexibility to describe the deviations. Both approaches extend the range of validity of the law but does not provide a solution general enough to properly characterize the absorption at any A value.

We propose to use \mv\ on a set of measurements to find such a general parametric solution. In order to proceed, several wavelength scans were carried out using the Hitachi double-beam spectrophotometer UH3500 from 800 nm to 200 nm. Thus, four molecules were analyzed at various concentrations in dichloromethane, a commercial coumarin, two bodipy \cite{NH2}\cite{COOM} and a porphyrin \cite{tpp} which was previously synthesized in order to measure their absorbance A as a function of the solution's concentration. Ideally, the parametric model should recover their extinction coefficient $\epsilon$.

After a small hyperparameter exploration we produce a simple accurate parametric function using \mv. It was produced with a max tree length of size 15 and the $exp$ and $log$ operators allowed.

\begin{equation}
    f(x; \mu, \epsilon) = \log\left(\frac{1}{\mu+\mathrm{e}^{-\epsilon x}}\right)
\label{eq:blb}
\end{equation}

\begin{figure*}
    \centering
    \includegraphics[width=0.7\textwidth]{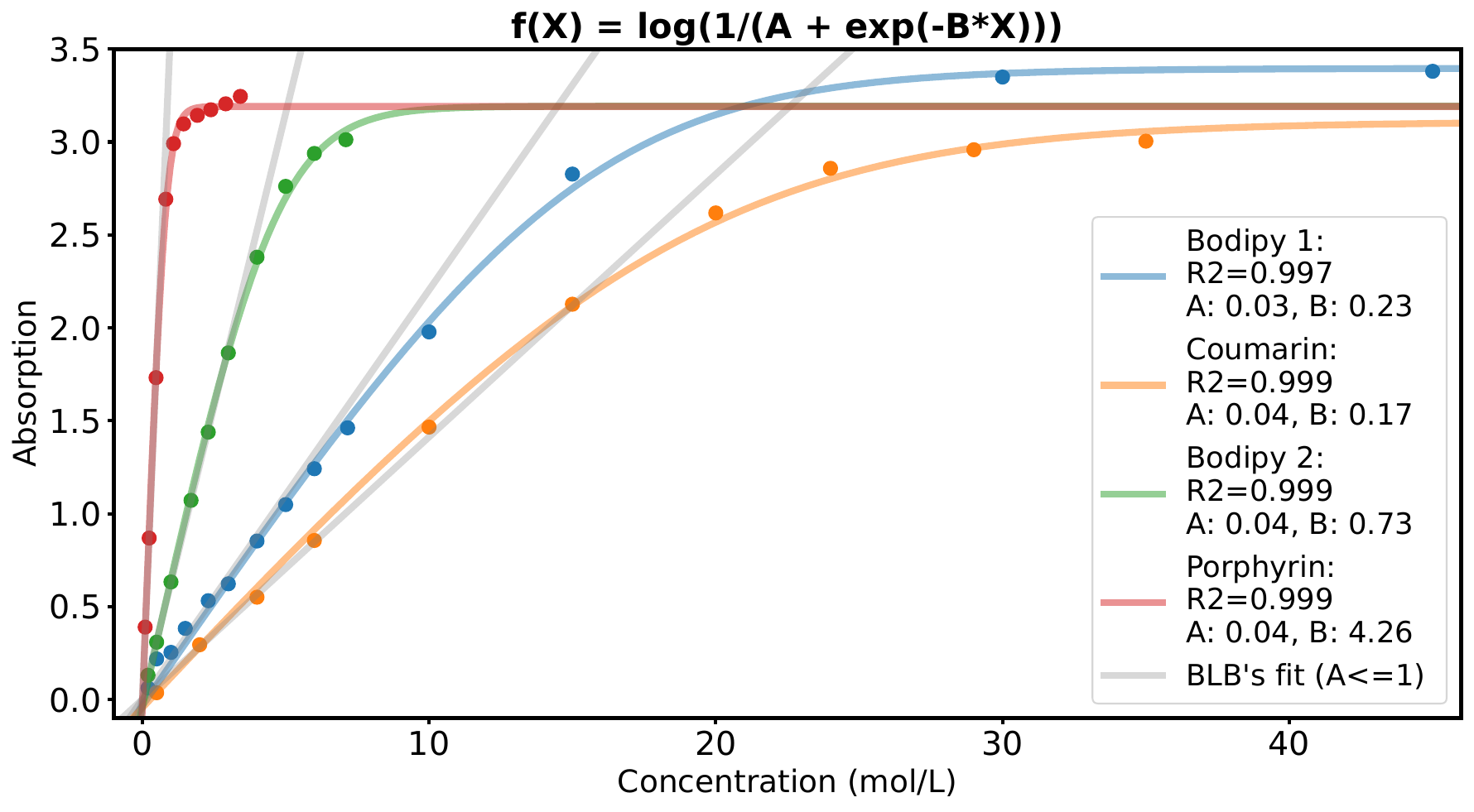}
    \caption{Best \mv\ fit (Equation \ref{eq:blb}) of the absorption as a function of the molar concentration for 4 different molecules. Gray lines correspond to the Beer's law fitted to the data points for which $A \leq 1$.}
    \label{fig:abso}
\end{figure*}

The data acquired display high non-linearity behaviors, in particular the absorption tend to reach a plateau around $A = 3$. Figure \ref{fig:abso} presents the best fits along with the parameter values associated and the $R^2$ score. The extended Beer's law proposed by \mv\ carries ideal properties to fit the observations. Indeed, in the case where the linearity is respected, the $\mu$ parameter can be set to 0 and the equation simplifies into the original Beer's law. In such case, $\epsilon$ carry exactly the same information as the standard method. In the general case, $1/\mu$ characterizes the plateau that the absorption will reach at high concentrations. \mv\ shows that an exponential transition between the linear evolution and the plateau provides an accurate fit to the data. We observe that all $R^2$ scores are close to 1, demonstrating that the law is fitted to the observations. Additionally, Figure \ref{fig:abso} displays the classical Beer's law fitted to the data points for which $A \leq 1$. We also observe that the lower the extinction coefficient is, the further apart the two models are at low absorption. This suggests a deviation from Beer-Lambert's law at low $\epsilon$, and would require further investigation.

In summary, Equation \ref{eq:blb} offers the possibility of computing molar extinction coefficients without being strictly restricted to the linear regime. The functional form proposed by \mv\ contains two parameters and thus requires only a few data points to be constrained. It could be 
an easiest and quickest alternative for the  determination of intrinsic properties of chemical species.

\subsection{Finance dataset}
\label{sec:economy}

Financial markets exhibit complex emergent behaviors. We define the return as the difference between the price of an asset at time $t$ and time $t+1$ is $r_t = p_{t+1} - p_{t}$. At first approximation, we can describe the distribution of returns by a \textit{Brownian motion}~\cite{bachelier1900theorie}. 
The field of econophysics, which applies tools from statistical physics to study these stochastic processes, seeks to identify probability distribution functions that go beyond such \textit{Gaussian} approximation. Those initial models led to the development of the renowned \textit{Black-Scholes} equation~\cite{Black1973}, but neglected the significance of rare and extreme events due to their lack of fat tails. Indeed, the distribution of these specific events can be fitted with \textit{power-laws}~\cite{Mantegna1995}.

Modern models, as presented in~\cite{mantegna1999introduction}, focus on \textit{Lévy processes}. These include, for example, the \textit{Gaussian} distribution as well as the \textit{Cauchy} distribution, the second being an already improved solution with its fat tails. Such models were proposed after studying the statistical properties that most data exhibit, such as the famous $S\&P500$ dataset, and then finding distributions that possess such properties. However, our Multiview approach enables us to identify a common distribution for all assets by considering each of them individually. 
In this section, we show that \mv\ rediscover some presented distributions and propose new models that better fit the real return distributions. 

We analyze time series data consisting of multiple prices generated by various assets. Each asset's price corresponds to the value of a share of a company on financial markets. We utilize a publicly available Kaggle dataset\,\footnote{\url{https://www.kaggle.com/datasets/iveeaten3223times/massive-yahoo-finance-dataset}} containing data from $491$ companies.
The first 10 companies are used as views for \mv, while the remaining companies are only used for testing purposes. Values of assets are taken each day at open market time, over a period of 5 years starting from January $1^{st}$, $2018$. For each asset, we study the distribution of its returns with a sampling of $100$ bins of equal width. We normalize the data as described in Section \ref{sec:data_gen}. This type of data exhibits common statistical properties. A positive mean that corresponds to overall economic growth. A leptokurtic profile that indicates more extreme events than a normal distribution would produce, and a negative skewness due to an asymmetric shape that informs rare events are more likely to be crises than economic booms. The first and last properties are only observed for datasets covering long time periods (months or years) such as ours. 

Under those conditions, we explore multiple seeds, tree lengths (ranging from size 8 to 20) and operators (using various combinations of $exp$, $abs$ and $power$) to obtain multiple possible parametric solutions with \mv. In Table \ref{tab:pdf_market} we present 6 parametric models generated by \mv. We measure the performance of all the functions by fitting them to each asset, and display the median MSE value as a comparative metric in the last two columns of Table \ref{tab:pdf_market}. 

In our experiment, \mv\ noticeably recovers 3 widely used parametric forms: the {Gaussian}, {Cauchy} and \textit{Laplace} distributions. In addition to these results, we obtain solutions similar to Laplace with some variations. These three distributions are presented in Table \ref{tab:pdf_market}. We can see from their scores that those parametric forms present better fits than the solutions already present in the literature. 

\begin{figure*}
    \centering
    \includegraphics[width=0.9\textwidth]{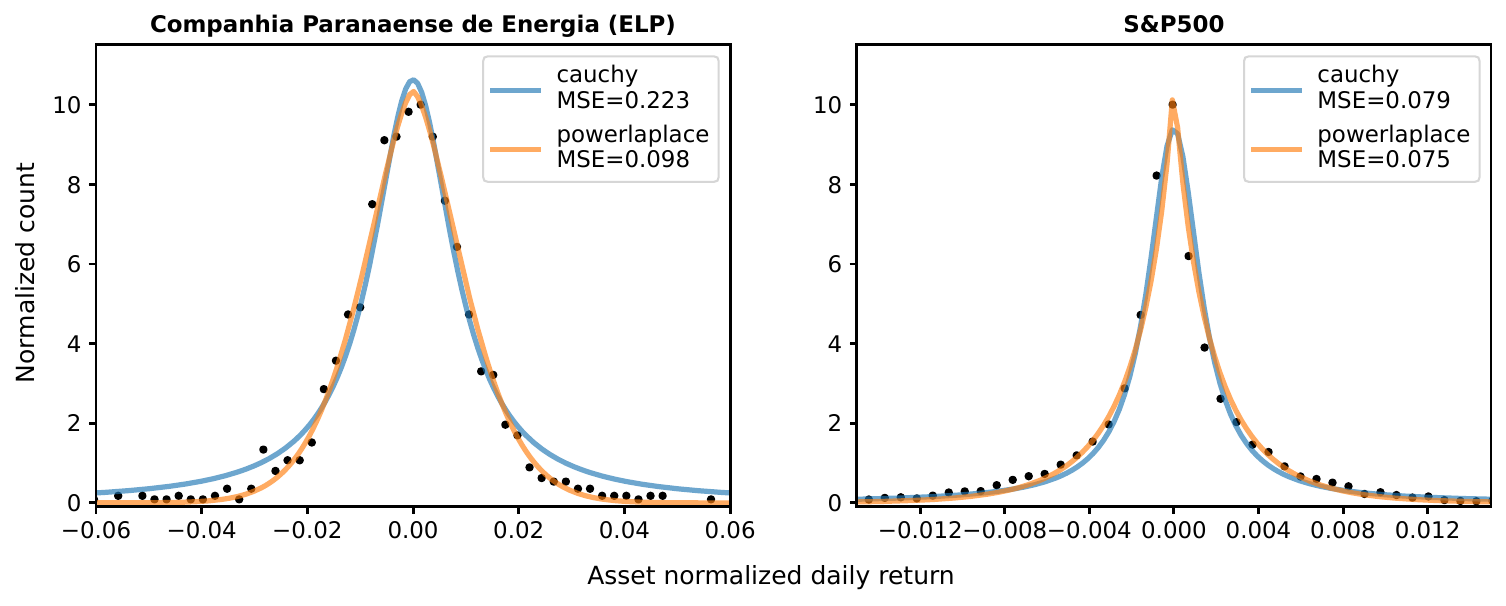}
    \caption{Normalized (Section \ref{sec:data_gen}) distribution of returns for 2 assets, fitted by the Cauchy model and the best \mv\ solution (Power-Laplace). The first asset is an example of a high MSE improvement in the usage of Power-Laplace compared to Cauchy.}
    \label{fig:market}
\end{figure*}

Figure \ref{fig:market} shows the distributions of returns of some assets alongside the fit given by the best distributions, new and already found in the literature, that we obtain. We call the latter a \textit{Power-Laplace} distribution\,\footnote{That we name after it's functional composition $h(x) = g(f(x))$ with the Laplace distribution $g(x) = e^{-a\times|x|}$ and a power function $f(x) = x^b$.}. It performs well because the power is always a root\,\footnote{We always find $0 < b < 1$.}, which accentuates the fat tail effect of the distribution, allowing a better fit in the tails while maintaining a good peaked fit in the center. 

\begin{table}[h!]
\centering
\begin{tabular}{|c||c|c|c|c|c|}

 \hline
   \textit{Models} & {Equation f(x)} & $med(MSE)$ & $MSE_{S\&P}$ \\ 
 \hline
 \hline 
  Gaussian~\cite{bachelier1900theorie, Black1973} & $A \cdot \mathrm{e}^{-\frac{x^2}{B}}$ & 0.363 & 0.260 \\ \hline
 Laplace~\cite{Kou2002} & $A\cdot \mathrm{e}^{-B|x|}$ & 0.342 & 0.084 \\
 \hline 
  Cauchy~\cite{Liu2012} & $A\cdot B^2 / (x^2 + B^2)$  & 0.305 & 0.079 \\ \hline 
 \hline
  Linear-Laplace & $(A - B x) \cdot \mathrm{e}^{-C |x|}$ & 0.327 & 0.065 \\
 \hline 
 Exp-Laplace & $A \cdot \mathrm{e}^{B x - C |x|} $  & 0.328  & \textbf{\textit{0.063}} \\ \hline
 Power-Laplace & $A \cdot \mathrm{e}^{B |x|^C}$  &  \textbf{\textit{0.246}} & 0.075 \\
 \hline  
 \end{tabular}
 \caption{Best functions generated by \mv. The last two columns respectively show the median MSE score of the functions fitted on individual normalized assets and the score when fitted on normalized the $S\&P500$ dataset. Bold numbers correspond to the best score of the column.}
 \label{tab:pdf_market}
\vspace{-8mm}
\end{table}

The last column of Table \ref{tab:pdf_market} shows the scores of all the functions fitted on the $S\&P500$ index. This index is an aggregated measure of the value of the assets of the 500 largest American companies. As such, the distribution of its returns should behave like an average distribution of these assets. For these data, the best distribution is the one in the second to last line of Table \ref{tab:pdf_market}. 
Like for the previous application on multiple assets, this shows our new distributions are also better fits than those present in the literature.

Finally, we see that distributions found in the literature are relevant for the $S\&P500$, as they all perform much better than the Gaussian. However, apart from the Gaussian, their scores do not differ much. This shows that they are almost all equivalent fits for this aggregated index. On the other hand, their median scores for a variety of assets are much more disparate. In this second case, our algorithm produced much more efficient distributions ; distributions that outperformed all the others. This indicates that our multiview approach is an efficient strategy for finding a unique distribution that gives good fits for as many assets as possible. We illustrate this effect on Figure~\ref{fig:market}. The right panel shows how Cauchy and Power-Laplace both fit well the $S\&P500$ index. The left panel displays a special case selected for its high score difference between both models. This shows how a model like Cauchy fails to fit some distributions compared to the best model found by \mv, which is more versatile.

We showed that applying \mv\ to a set of assets all at once, rather than looking at them separately or using an aggregated index, is an efficient approach to gaining new insights into markets behavior and improving characterization of the stochastic process that governs them. The results presented here could be fine-tuned by adding more functions to the \mv. For example, the \textit{Gamma function} could allow us to generate more distributions 
such as the \textit{Gamma variance process}~\cite{Madan1998} and \textit{Student's t-distribution}~\cite{bouchaud2000theory}, which are the general forms for the distributions we already found using \mv. 
Finally, further investigation of the distributions found here may prove insightful for future modelling in the field.

\subsection{Astrophysics dataset}
\label{sec:astro}

Astrophysics constitutes an ideal testing environment for \mv\ as it encompasses a diverse array of non-linear equations aimed at modeling observations of large-scale phenomenon within the universe. Classical SR has been widely used to characterize various relationships such as photometric redshift\cite{IshidaSR}, the Hertzsprung–Russell diagram or the plane of elliptical galaxies \cite{GrahamSR}. After this project was completed, we also  became aware of \cite{fakemvsr}, who approached a similar problem when dealing with the treatment of stellar streams.

Astrophysical phenomena present a diverse landscape due to the vast range of physical values in stellar objects and the variety of physical processes occurring within them. One method to study these phenomena involves tracking the change in radiation flux over time. This time-series of flux measurements, known as a light curve, typically varies in the number of observations depending on the object and the optical filter in use. This is particularly true for ground-based surveys, such as the Zwicky Transient Facility (ZTF,~\citep{ztf2019}), because of weather conditions, seasons and other factors.

The largest catalogs of variability classifications contain millions of stellar objects~\citep{vsx,gaia_dr3_variables}, providing researchers access to numerous light curves of the same type of variability. 
One of the most studied types of extreme variability is supernovae (SNe), a catastrophic cosmic event caused by a star’s explosion. The physical processes in a supernova are highly complex, and we may not yet be able to directly model individual supernova light curves. This makes such astrophysical data an excellent case study for \mv.

We choose three Supernova type Ia (SNIa) with a good time sampling from ZTF Data Release 17: SN2019fck, SN2018aye and SN2021mwb \citep{ztf2019,snad_viewer}.
We use observations in $g$ and $r$ photometric filters independently, resulting in a dataset of six examples for \mv.
The light curve shape of SNIa in $g$ filter could be roughly described as an exponentially rising brightness followed by an exponential fading.
Such behavior is typically well modeled by  parametric equations found in the literature~\citep{bazin,Villar_2019,alerce_lc_classifier,rainbow}.
However, being observed in a redder $r$ filter, SNIa light curves have a secondary bump. This particular behavior is much harder to describe with a simple parametric equation.
We choose to use both $g$ and $r$ band to provide a wide variety of examples to \mv.

We apply data quality cuts and select only data points with a signal-to-noise ratio greater than 20. The data was shifted so that the observed peak time is at 0 and the entire light curve was normalized by its maximum flux. Finally, we consider only data points with a time ranging from -50 to 150 days from the observed peak.

\begin{table}[t!]
\centering
\begin{tabular}{|c|c|c|}
 \hline
 {Equation f(t)} & $<R^2>$ & No. parameters\\
 \hline
 \hline 
 $\mathrm{e}^{-At \cdot (B - \mathrm{e}^{-Ct})}$ & 0.990 & 3 \\ 
 \hline
 $\frac{A}{(B\cdot \mathrm{e}^{Ct}+\mathrm{e}^{-Dt})}$ & 0.987 & 4 \\
 \hline 
 $\frac{A^{Bt}}{Ct + (-Dt + \mathrm{e}^{Et})^2}$  & 0.992 & 5 \\ 
 \hline 
 \end{tabular}
 \caption{Summary of the best parametric functions generated using \mv\ on SNIa lightcurves. The second column corresponds to the mean $R^2$ score over the 6 examples provided.}
 \label{tab:astro_result}
 \vspace{-4mm}
\end{table}

We use a maximum tree length of 12, include the $exp$, $square$ and $power$ operators and explore multiple seeds to obtain a panel of possible parametric solutions. Table \ref{tab:astro_result} displays a selection of the best parametric solutions found by \mv\ along with the mean $R^2$ score of the best fit on the SNe. 

\begin{figure*}
    \centering
    \includegraphics[width=0.9\textwidth]{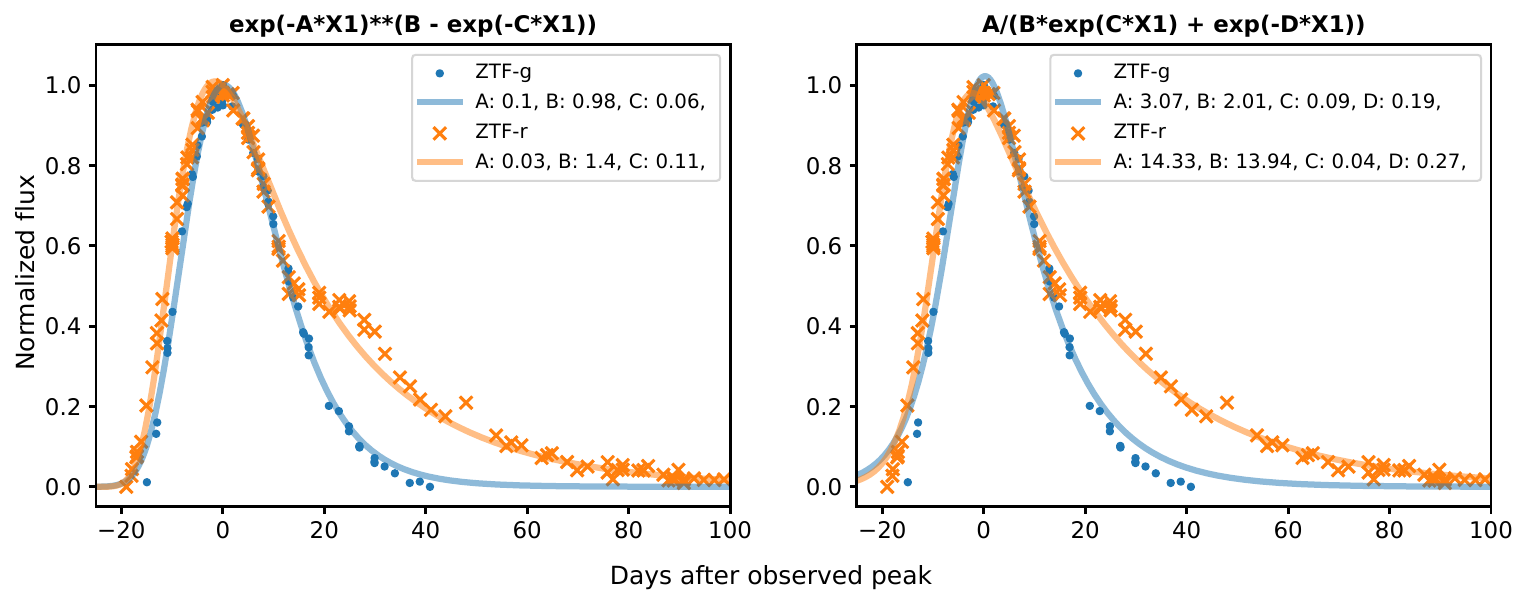}
    \caption{Best fit of two parametric functions found by \mv\ on SN2021mwb in the $g$ and $r$ filters. The right panel corresponds to the Bazin function commonly used in the literature.}
    \label{fig:snia}
\end{figure*}

Noticeably, the second equation is already known as the \
Bazin function \cite{bazin} and is widely used in the literature \cite{rainbow}\cite{bazin_ex1}. However, the standard form includes a \textit{$t_0$} parameter appearing twice which is added to \textit{t}, effectively encoding for a time shift. The form generated by \mv\ is mathematically equivalent but uses parameters appearing only once. The first solution presented in Table \ref{tab:astro_result} is characterized by an intricate exponential form. Despite lacking similar counterparts in the literature, it provides excellent fits and even outperforms the standard Bazin function (Figure \ref{fig:snia}). Moreover, since it uses only 3 parameters, additional ones such as \textit{$t_0$} or a scaling parameter could be added to make it more versatile. Finally, the last equation in Table \ref{tab:astro_result} requires 5 parameters and provides the most accurate description of the SNe. \mv\ can produce solutions of arbitrary size with increasingly good fit, however we choose to limit it to 5 parameters maximum in order to prevent the overfitting.

Overall, \mv\ was able to generate multiple good models to describe SNe behaviors. It recovered a solution from the literature and even proposed improved solutions in terms of goodness of fit and/or number of parameters used. However, the generated models struggle with the same problem as the equations from the literature: they don’t provide a simple description of the second bump of the SNIa in the $r$ band. It may also highlight a limitation of the current \mv\ implementation, which doesn't include parameter repetition, as stated in Section \ref{sec:description}). Given that a repeated time shift parameter is standard in models used in the literature, this results highlights the importance of a complete \mv\ implementation for practical scientific applications.

\vspace{0.3cm}
The code used to produce all the results presented in this work is publicly available\footnote{\url{https://github.com/erusseil/MvSR-analysis}}.

\section{Conclusions}
\label{sec:conclu}

Symbolic Regression (SR) has proven to be extremely efficient in searching for mathematical expressions that describe the relationship between a set of explanatory and response variables. In its traditional form, it translates the behavior of one such data set into an analytical function which can be used for further analysis. Nevertheless, in a realistic scenario, the researcher is frequently faced with multiple outcomes from the same experiment. These may correspond to different experimental setups, initial conditions or domain coverage, but are all generated by the same underlying mechanism, one which the researcher aims to describe.

In this work, we proposed Multi-View Symbolic Regression (\mv), a framework that exploits this scenario by extending the scope of traditional SR, allowing the user to provide multiple examples as input. The algorithm searches for the best parametric function which simultaneously describes all the input data provided. This is achieved by fitting each input data set independently with the same regression model and aggregating their individual fitness into a single one.

For this purpose, we included an aggregation fitness function in Operon that supports different aggregators such as mean, maximum, median, or harmonic mean. 

We have tested this approach with four different challenging artificial benchmarks with added noise and composed of extreme situations where some parameters are set to $0$ or the domain coverage is limited. When compared to the traditional approach, we report that \mv\ is capable of correctly retrieving the original expression in most configurations with a higher accuracy than SR, even in a presence of a strong noise. Additionally, we stress-tested our method by using real-world experiments from three different areas: chemistry, finance and astrophysics. \mv\ was not only capable of recovering well known models from the literature, but it also found new alternatives that are promising in these fields.

Our results showcase the potential enclosed in applying  \mv\ in real scientific scenarios. These could be made even better with further functionalities, like enabling a maximum number of parameters to be used in the model, either as a hard constraint or through a penalization term, and allowing the same parameter to appear more than once in the final expression. Such additional features would not only result in more flexible functions, with smaller number of parameters, but it would also allow the researcher to indirectly tailor the final result, thus increasing the chances of a parametric form which can inspire interpretability. 

In this context, a full implementation of \mv\ would have much broader applications than the ones described here. We intend to further explore its ramifications in dedicated studies focused on the particularities of each science case. Nevertheless, in  order to fully exploit its potential and popularize its use within a broad range of scientific areas, a user-friendly implementation of all the above-mentioned functionalities is paramount.

\section*{Acknowledgements}

This project emerged from discussions during two SNAD workshops: SNAD-IV\footnote{\url{https://snad.space/2021/}} and SNAD V\footnote{\url{https://snad.space/2022/}}, held in France,  Brian\c{c}on and Clermont Ferrand, respectively. ER, KM and EEOI thank the entire SNAD team for the insightful discussions and environment. ER acknowledges the ISITE grant from the "AAP Séjour-recherche" program from Universit\'e Clermont-Auvergne (UCA). FOF acknowledges support from Funda\c{c}\~{a}o de Amparo \`{a} Pesquisa do Estado de S\~{a}o Paulo (FAPESP), grant number 2021/12706-1 and CNPq through the grant 301596/2022-0. EEOI thanks Rafael S. de Souza and Alberto Krone-Martins for the introduction to Symbolic Regression. Finally we thank the Brazilian past\'eis for being a true source of inspiration. 

\bibliographystyle{ACM-Reference-Format}
\bibliography{sample-base}

\end{document}